\documentclass[10pt,twocolumn,letterpaper]{article}

\usepackage{iccv}
\usepackage{times}
\usepackage{epsfig}
\usepackage{graphicx}
\usepackage{amsmath}
\usepackage{amssymb}

\usepackage[ruled,vlined]{algorithm2e}
\usepackage{ctable}
\usepackage{caption}
\usepackage{enumitem}

\renewcommand{\paragraph}[1]{\vspace{1mm}\noindent\textbf{#1}}
\newcommand{\Tref}[1]{Table~\ref{#1}}

\newcommand{\fref}[1]{Fig.~\ref{#1}}

\newcommand{\sref}[1]{Sec.~\ref{#1}}

\newcommand{\aref}[1]{Alg.~\ref{#1}}

\newcolumntype{P}[1]{>{\centering\arraybackslash}p{#1}}

\usepackage[pagebackref=true,breaklinks=true,letterpaper=true,colorlinks,bookmarks=false]{hyperref}

\iccvfinalcopy 

\def\iccvPaperID{2840} 

\ificcvfinal\fi
\begin{document}

\title{Onion-Peel Networks for Deep Video Completion}

\author{
Seoung Wug Oh\\Yonsei University \and Sungho Lee\\Yonsei University \and Joon-Young Lee\\Adobe Research \and  Seon Joo Kim\\Yonsei University
}

\maketitle
\thispagestyle{empty}

\begin{abstract}
We propose the onion-peel networks for video completion. 
Given a set of reference images and a target image with holes, our network fills the hole by referring the contents in the reference images.
Our onion-peel network progressively fills the hole from the hole boundary enabling it to exploit richer contextual information for the missing regions every step.
Given a sufficient number of recurrences, even a large hole can be inpainted successfully.
To attend to the missing information visible in the reference images, we propose an asymmetric attention block that computes similarities between the hole boundary pixels in the target and the non-hole pixels in the references in a non-local manner.
With our attention block, our network can have an unlimited spatial-temporal window size and fill the holes with globally coherent contents.  
In addition, our framework is applicable to the image completion guided by the reference images without any modification, which is difficult to do with the previous methods.
We validate that our method produces visually pleasing image and video inpainting results in realistic test cases.
\end{abstract}

\section{Introduction}

\begin{figure}
\centering
\includegraphics[width=1.0\linewidth]{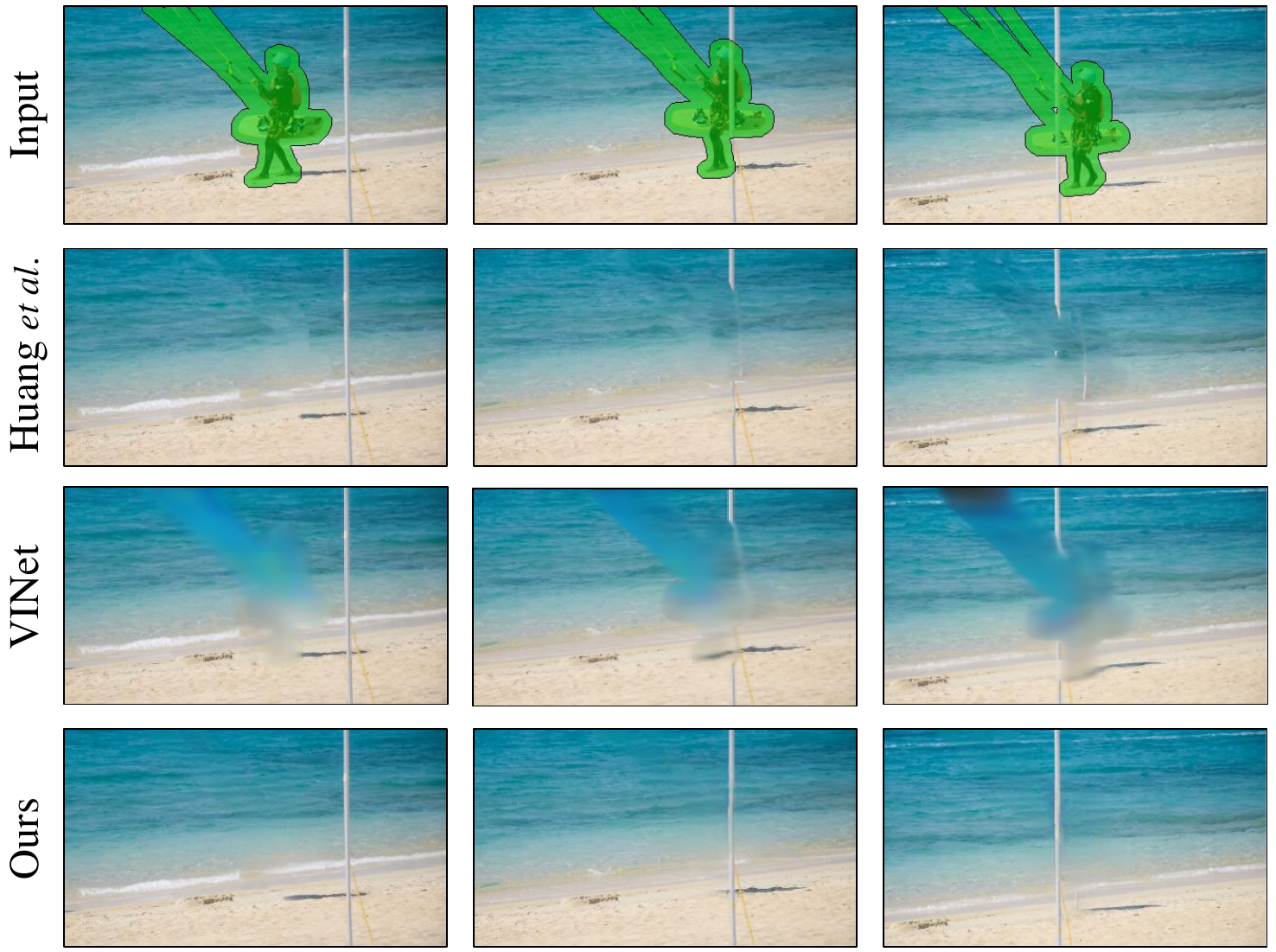}
\caption{Video completion results from  Huang~\etal~\cite{Huang2016siggraph}, VINet~\cite{kim2019deep}, and our method. Our method is able to handle challenging cases where the flow computation is not easy due to occlusions or large holes.}
\label{Fig:teaser}
\end{figure}

Recently, we have seen rapid progress in the image completion task -- synthesizing missing pixels in an image -- using deep neural networks~\cite{IizukaSIGGRAPH2017, yu2018generative, PathakCVPR2016, Liu2018PartialConv, yu2018free}.
Generative adversarial networks~\cite{goodfellow2014generative} are often used to \textit{hallucinate} missing contents and have shown impressive results.
Despite the progress in image completion, video completion has been rarely explored using deep networks and still remains in an early stage.
In video completion, we need to fill missing regions with coherent contents through time rather than synthesizing contents at each frame independently. 
When removing an object in a video, the region occluded by the object may be visible in other (potentially distant) frames, and filling the region without considering the original content in other frames will severely break the temporal consistency.
Therefore, it is difficult to directly extent ideas of image inpainting networks to the video completion problem using 3D convolutional or recurrent networks because their temporal receptive fields are too limited or directional.

For video completion, the traditional approach like the global flow field based optimization technique~\cite{Huang2016siggraph} produces outputs with state-of-the-art quality. Although it performs well in many cases, it is slow because it is computationally intensive and difficult to parallelize. In addition, it suffers from challenging cases where optical flows are noisy (\eg occlusions or large holes).
Recently, two deep learning based video completion methods have been proposed. 
CombCN~\cite{wang2018videoinp} is built by combining 2D and 3D CNNs, but it is only validated on low-resolution videos of aligned faces and vehicles with fixed square holes, making it difficult to be used for real object removal cases.
VINet~\cite{kim2019deep} is designed as a recurrent network and internally computes the flow fields from 5 adjacent frames\footnote{[$t$-6, $t$-3, $t$-1, $t$+3, $t$+6], where the current frame is at $t$} to the target frame. 
Due to the network design, its temporal window is restricted to the nearby frames, therefore, it is difficult to inpaint holes with the faithful content that can be visible outside the temporal window. 
Both works are meaningful in tackling the challenging video completion task using deep networks and produce video inpainting results much faster than the traditional methods, however, they hardly improve the output quality over the traditional methods.

In this paper, we propose a novel deep network called Onion-Peel Network (OPN) for video completion. 
Given video frames with inpainting masks, we use a set of (sampled) frames as the reference and fill the hole of the target frame by taking the contents from the reference frames, or synthesizing coherent contents if there is no missing content visible in the reference.
Our network inpaints the hole region one layer (peel) at a time by gradually eroding the hole ~\cite{Newson2014siam, newson2017non}\footnote{The term "onion peel" originated from the initialization technique in~~\cite{Newson2014siam, newson2017non} for the randomized search through the PatchMatch~\cite{barnes2009patchmatch}}.  
By doing so, our network can exploit richer contextual information for the missing regions at each step. 
Given a sufficient number of recurrences, even a large hole can be inpainted successfully.

To pick up the missing contents that are visible in the reference images, we propose an asymmetric attention block that computes similarities between the peel pixels in the target and the non-hole pixels in the references in a non-local manner.
With our attention block, our network can have an unlimited spatial-temporal window to the reference and inpaint holes with globally coherent contents.  
Our attention-based approach is also free from optical flow computations, thus we are able to handle challenging scenes with occlusions or large holes hardly modeled by flows.


In addition to video completion, our network is also applicable to image completion with reference images. 
Unlike the single-image inpainting task, we take a target image with additional reference images as input and we aim to complete the target image faithfully to the reference.
It is useful for image editing as people tend to take multiple photographs of a scene in different angles and time. 
Using a group of photos, our method enables one to remove undesired objects without damaging the original contents.
While this scenario can be considered as a special case of video completion with only a few frames, previous video completion methods have difficulties in handling this because computing optical flow between distant frames is challenging.

To summarize, we make the following contributions for accurate video completion:
\begin{itemize}[nosep,itemsep=0pt,leftmargin=10pt]
    \item[--] We propose the novel onion-peel network for flow-free video completion that works by the spatio-temporal attention mechanism.
    \item[--] We propose the asymmetric attention block that allows non-local matching between the hole and the non-hole pixels. 
    \item[--] Our framework is applicable to the image completion guided by the reference images that is hard to be achieved by the previous image and video completion methods. 
    \item[--] We validate that our method produces comparable results to the state-of-the-art methods with fast computational time.
\end{itemize}

\begin{figure*}
\centering
\includegraphics[width=1.0\linewidth]{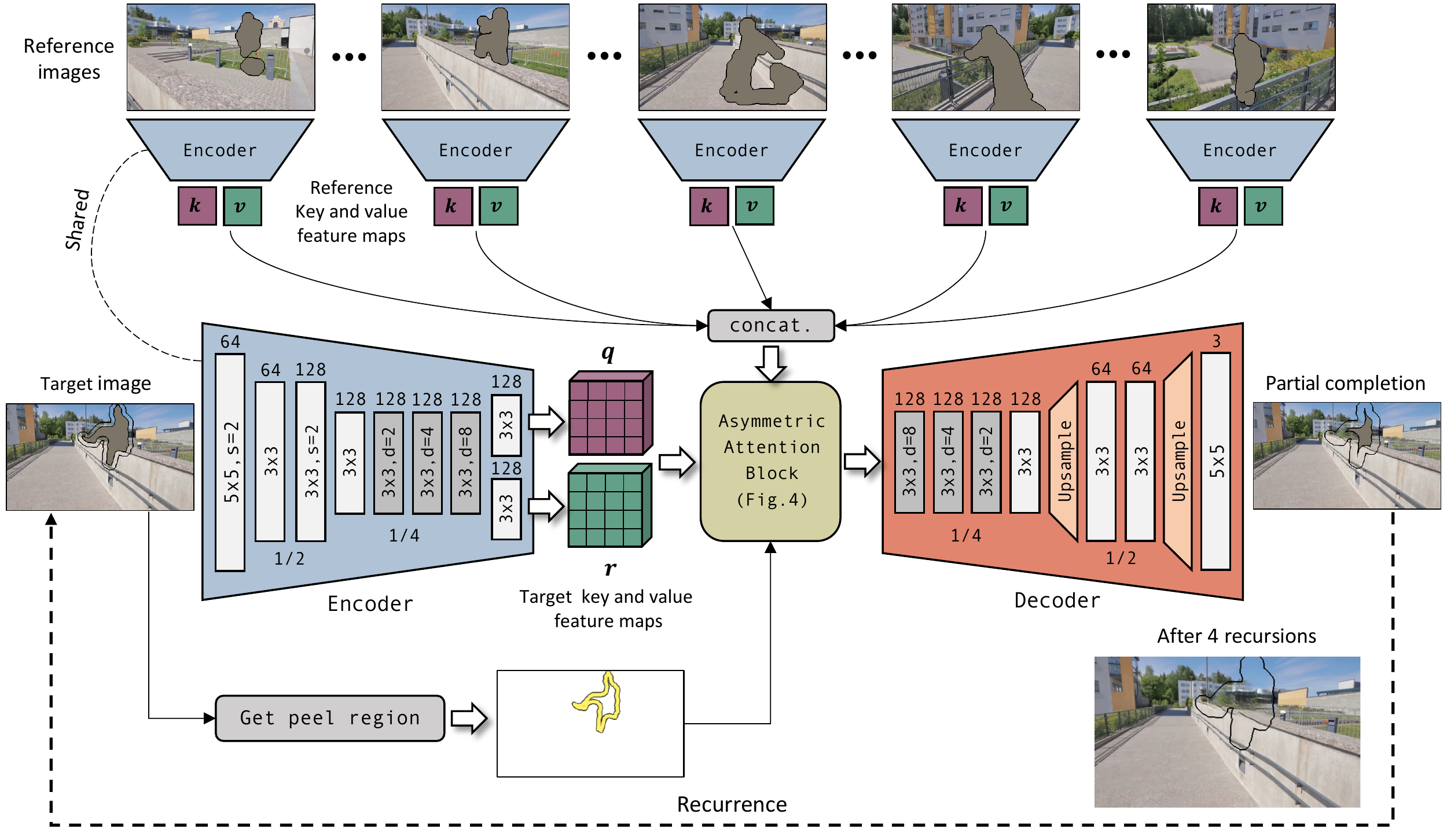}
\caption{Overview of our network. We use the gated convolution layer as a basic building block. $s$ and $d$ indicate the stride and the filter dilation of the gated convolution layer. If not indicated, we set $s$=1 and $d$=1.}
\label{Fig:overview}
\end{figure*}

\section{Related Work}
Given the usefulness and difficulty of the problem, there is a large body of literature on the image and video completion. 
We review representative and recent studies that are closely related to our method. 
We refer the survey papers -- image~\cite{guillemot2014image} and video~\cite{ilan2015survey} completion -- for a comprehensive review.

\subsection{Image Completion.}
\paragraph{Example-based methods.}
Example-based methods rely on the self-examples to inpaint the missing regions~\cite{criminisi2004region, hays2007scene, liu2013exemplar}.  
In specific, a missing region is reconstructed by referencing the information outside the missing region.
Furthermore, PatchMatch~\cite{barnes2009patchmatch}, a randomized search method for a fast approximation of the nearest neighbor field, accelerates the inpainting for the high quality results~\cite{newson2017non, huang2014image}. 
A commercial solution in Adobe Photoshop belongs to this category. 

\paragraph{Learning-based Methods.}
For learning-based methods, deep generative models are trained to synthesize missing contents in a data-driven manner. 
Generative adversarial networks~\cite{goodfellow2014generative} are often adopted for this purpose~\cite{PathakCVPR2016, yu2018generative, IizukaSIGGRAPH2017, yu2018free}.
To specifically handle incomplete data in the inpainting problem, the partial~\cite{Liu2018PartialConv} and gated convolution~\cite{yu2018free} were proposed as an alternative to the normal convolution operator.
We use the gated convolution layer in our network.

In this work, we tackles the multi-image completion that shares the similar concept of the example-based methods.
The difference is that we find the example information from the reference images rather than from the same image.

\subsection{Video Completion}
\paragraph{Patch-based Methods.}
Wexler~\etal\cite{wexler2004space} present a global optimization approach that alternates between the nearest neighbor search for the 3D spatio-temporal patches and the pixel reconstruction. 
Newson~\etal\cite{Newson2014siam} extends \cite{wexler2004space} by employing the PatchMatch~\cite{barnes2009patchmatch} to accelerate the nearest neighbor search. 
Huang~\etal\cite{Huang2016siggraph} presented the non-parametric optimization method based on the dense optical flow field. 
The authors combine the spatial patch-based optimization and the flow field estimation to achieve spatially and temporally coherent inpainting. 
While this method is recognized as the state-of-the-art, the execution time is extremely long due to the computational complexity.   

Compared to the above methods that perform the global optimization, our method runs an order of magnitude faster as it is based on feed-forward neural network. 
In addition, our data-driven approach is more flexible in handling unseen backgrounds.

\paragraph{Data-driven Methods.}
Recently, data-driven methods for video completion through the deep networks have been proposed~\cite{wang2018videoinp, kim2019deep}. 
In~\cite{wang2018videoinp}, the authors propose to use a combination of 3D and 2D CNNs, where the 3D sub-network is working with low-resolution video to reduce the computation. 
In~\cite{kim2019deep}, an image-based encoder-decoder network is designed to collect information from the nearby frames. 
Specifically, a set of the optical flow is estimated to aggregate information from the neighbor frames.

As mentioned earlier, those approaches have a limitation on their spatial-temporal window, which is crucial for the global coherency.  
Therefore, these methods are not expected to produce plausible results when a wide temporal window is required (\eg, the object is big or moving slowly).
On the other hand, our approach exploits a non-local pixel matching to have a global temporal window, thus it is free from such problems.

\section{Onion-peel Network}
The overview of our network for video completion is shown in \fref{Fig:overview}.
Given each video frame annotated with regions to be filled, the goal is to fill the hole region by looking at the other frames in the video for the right pixels.
We call the image to be filled as the \textit{target} image and the other images as the \textit{reference} images.
At each step, the onion-peel network fills only the \textit{peel} region of the target image by referring to valid regions on the reference images.
The peel region is defined as the set of pixels that are on the boundary of the hole region. 
We process the target image recursively to gradually fill the hole using the reference images.

\begin{figure}
\centering
\includegraphics[width=0.8\linewidth]{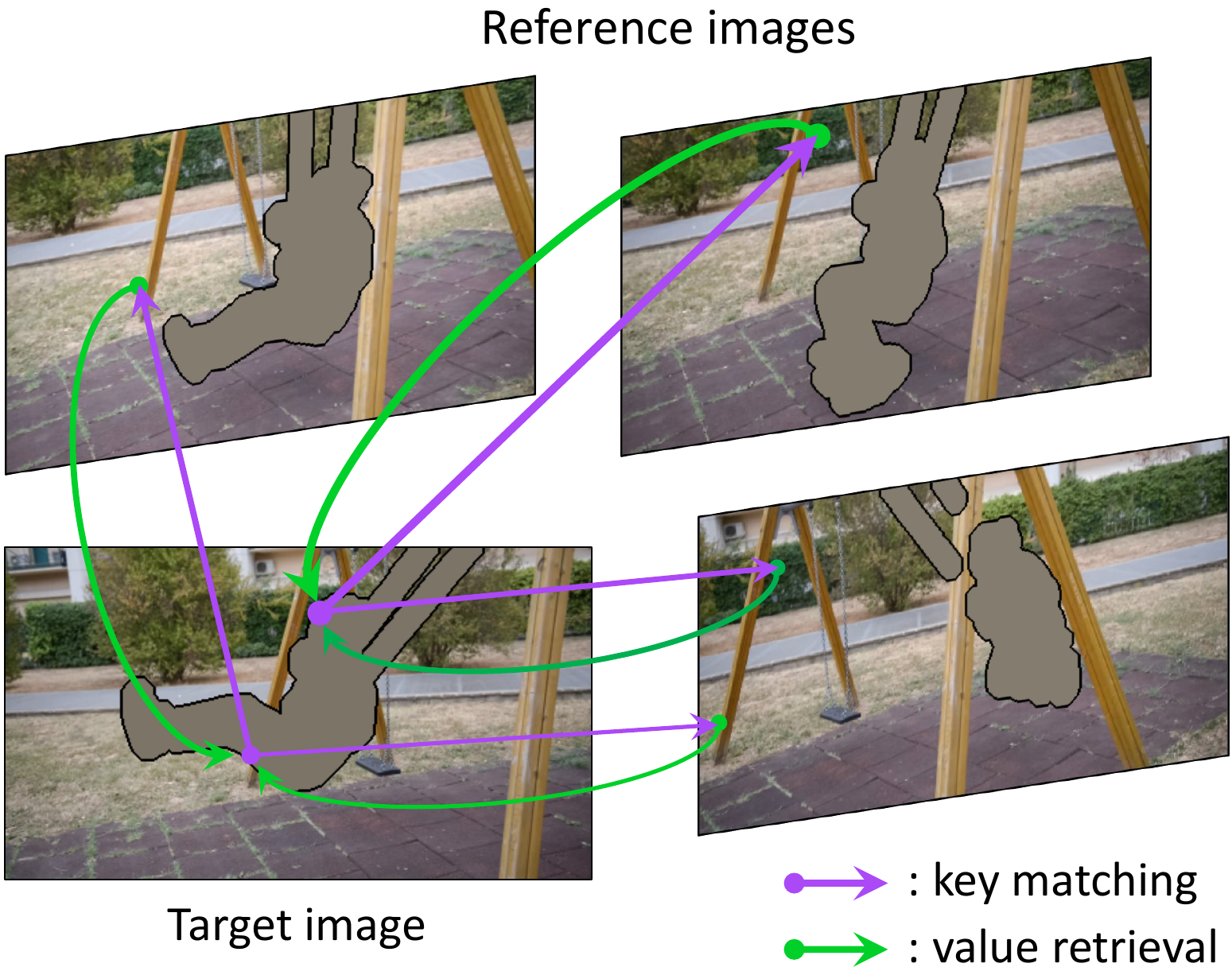}
\caption{A visualization of attentions computed in our asymmetric attention block. The thickness of arrows represents the matching (softmax) score.}
\label{Fig:attention}
\end{figure}

In our network, the target image as well as the reference images are first embedded into the \textit{key} and the \textit{value} feature maps through the shared encoder network.
We use the key features to find correspondences between the pixels in the target and the reference images.
The key features on the peel region in the target image will be matched to every key feature on all valid (non-hole) regions of the reference images using the asymmetric attention block.
The result of the matching is a spatio-temporal attention map, giving information about which pixel in which frame is important to fill the pixels in the peel region. 
The value is an abstract feature representation of an image to be used for reconstructing output images. 
Using the attention map from the key matching, we retrieve value features on the non-hole region of the reference frames according to the matching score.
An example of the key matching and the value retrieval is shown in~\fref{Fig:attention}.
The retrieved value features from the reference feature map are added to corresponding value features on the peel region in the target feature map. 
Then the decoder takes the updated target value feature map with the peel region mask and inpaints the peel region of the target image.

\subsection{Inference}
We present two applications of the proposed onion-peel network: reference-guided image completion and video completion.
For the image completion, reference images are embedded into key and value feature maps through the encoder. 
Then, the whole network (encoder, asymmetric attention block, and decoder) is applied to the target image recursively to fill the peel region iteratively until the hole is completely filled.
For the video completion, the procedure of the image inpainting is looped over every frame sequentially. 
In the case of video, a set of the reference images is sampled from the video. 
In our implementation, we sampled every 5-th frame as the references.

The detailed procedure of the video inpainting is described in~\aref{Alg:onion}. 
Image completion is a specific case of the video completion with one target image and a set of reference images.  
Here, we define $i$-th video frame as $X_i$ and its hole mask as $H_i$. 
The peel region $P$ is defined as the hole pixels that are within the Euclidean distance of $p$ to the nearest non-hole pixel.
We set $p$ to 8. 
The validity map $V_i$ indicates genuine non-hole pixels that are not filled by the algorithm. Because the reference frames are not changed during inpainting a video, we run the encoder for the reference frames only once.

\begin{algorithm}
\DontPrintSemicolon
\SetAlgoLined
\SetKwInOut{Input}{Inputs}\SetKwInOut{Output}{Output}
\Input{video $X$, hole $H$, validity $V$, peel width $p$}
\Output{completion video $X$}
\For{l $\textnormal{\textbf{in}}$ reference frame indices}{
    $\mathbf{k}_l, \mathbf{v}_l = \text{Encoder}(X_l, H_l, V_l)$\;
    }
\For{i $\textnormal{\textbf{in}}$ target frame indices}{
    $X_i^0 = X_i$\;
    $H_i^0 = H_i$\;
    $j = 0$\;
    \While{$\|H_i^j\|$}{ 
    $P^j = \text{get\_peel}(H_i^j, p)$\; 
    $\mathbf{q}, \mathbf{r} = \text{Encoder}(X_i^j, H_i^j, V_i)$\; 
    $\mathbf{z} = \text{AsymAtteBlk}(\mathbf{q}, \mathbf{r}, P^j, \mathbf{k}_{\forall\setminus i}, \mathbf{v}_{\forall\setminus i}, V_{\forall\setminus i})$ \;
    $\hat{X}_i^{j+1} = \text{Decoder}(\mathbf{z})$\;
    $X_i^{j+1} = (1-P^j)\odot X_i^j + P^j \odot \hat{X}_i^{j+1} $ \;
    $H_i^{j+1} = H_i^j  - P^j $ \;
    $j = j + 1$\;
    }
}
\caption{Onion-peel hole filling.}
\label{Alg:onion}
\end{algorithm}

\subsection{Network Design}
\paragraph{Encoder.} 
The input to the encoder network consists of an RGB image, the hole mask, and the validity map. 
The hole pixels on the RGB image are filled with the neutral (grey) values. 
These inputs are concatenated along the channel axis to form a 5-channel image before being fed into the first layer. 
We use the gated convolutional layer as a basic building block as it is useful for handling void information like a hole~\cite{yu2018free}.
The encoder downsamples the feature map up to the 1/4 scale of the original size to secure the high-frequency details. 
The dilated convolution is used to further enlarge the receptive field size, and similar design choices can be found in recent image completion works~\cite{IizukaSIGGRAPH2017, yu2018generative}. 
The encoder has two output heads each for the key and the value embedding, and this is implemented as two parallel gated convolutional layers.

The same encoder network is used for both the target image and the reference images to embed them into the key and the value feature maps.
We denote the key and the value map from the target frame as $\mathbf{q}$ and $\mathbf{r}$, and the key and the value maps from the reference frames as $\mathbf{k}$ and  $\mathbf{v}$.

\begin{figure}
\centering
\includegraphics[width=1.0\linewidth]{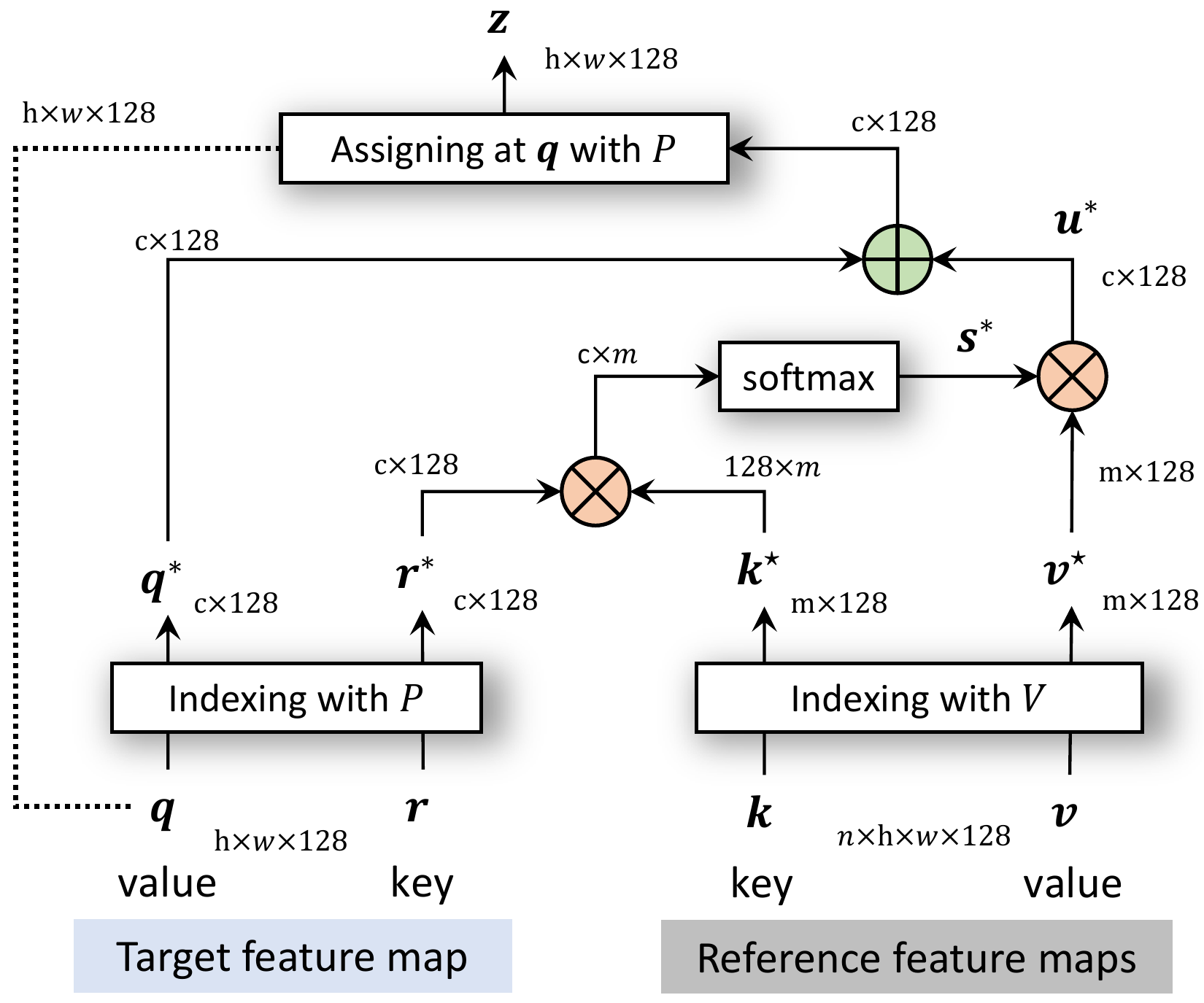}
\caption{Asymmetric attention block. Variables after indexing are marked with high asterisks. $h$ and $w$ are the spatial resolution of feature maps, and $n$ is the number of the reference frame. $c$ and $m$ is the number of pixels on the peel region and non-hole regions. $\bigotimes$ and $\bigoplus$ represent the matrix multiplication and element-wise addition, respectively.}
\label{Fig:read}
\end{figure}

\paragraph{Asymmetric Attention Block.} 
The key and the value feature maps from the target image and the reference images are fed into the asymmetric attention block. 
A detailed illustration is shown in~\fref{Fig:read}.
In this module, inspired by recent attention mechanisms~\cite{sukhbaatar2015end,wang2018non}, the key feature maps from the target and the reference are non-locally matched to compute a soft attention map over the reference images.
Then, the attention map is used for the retrieval of the reference value features. 
As the network aims to reconstruct the peel region of the target image by looking the non-hole region of the reference images, we constrain the matching targets accordingly using the peel $P$ and the validity map $V$.
Specifically, before the key matching, pixels on the target and the reference embedding maps are indexed by $P$ and $V$, respectively.
This results in smaller matrices (marked with high asterisks in \fref{Fig:read}) where the column size is the number of pixels on the peel and the valid region, respectively. 
By doing so, we can specify the region of interests and the computational cost is significantly reduced as the key matching and value retrieval are implemented as a matrix inner-product.
The matching scores $\mathbf{s}^*$ are cosine similarities normalized by the softmax function.
The reference value features are then retrieved by the weighted summation with the computed matching scores.
The retrieved values $\mathbf{u}^*$ are added to the target values $\mathbf{q}^*$ and assigned back to the original peel position ($P$).
An example of this attention mechanism is visualized in~\fref{Fig:attention}.

\paragraph{Decoder.}
The decoder takes the output of the asymmetric attention block to reconstruct the peel region of the target image. 
We make the decoder to focus on the recovery of the peel region through our loss function which put more weight on that area (see~\sref{Loss} for detail). 
The decoder network is designed to be symmetric to the encoder network.
The nearest neighbor upsampling is used to enlarge the feature map and the peel region of raw decoder output is combined with the non-hole region of the input target frame.

\begin{figure}
\centering
\includegraphics[width=1.0\linewidth]{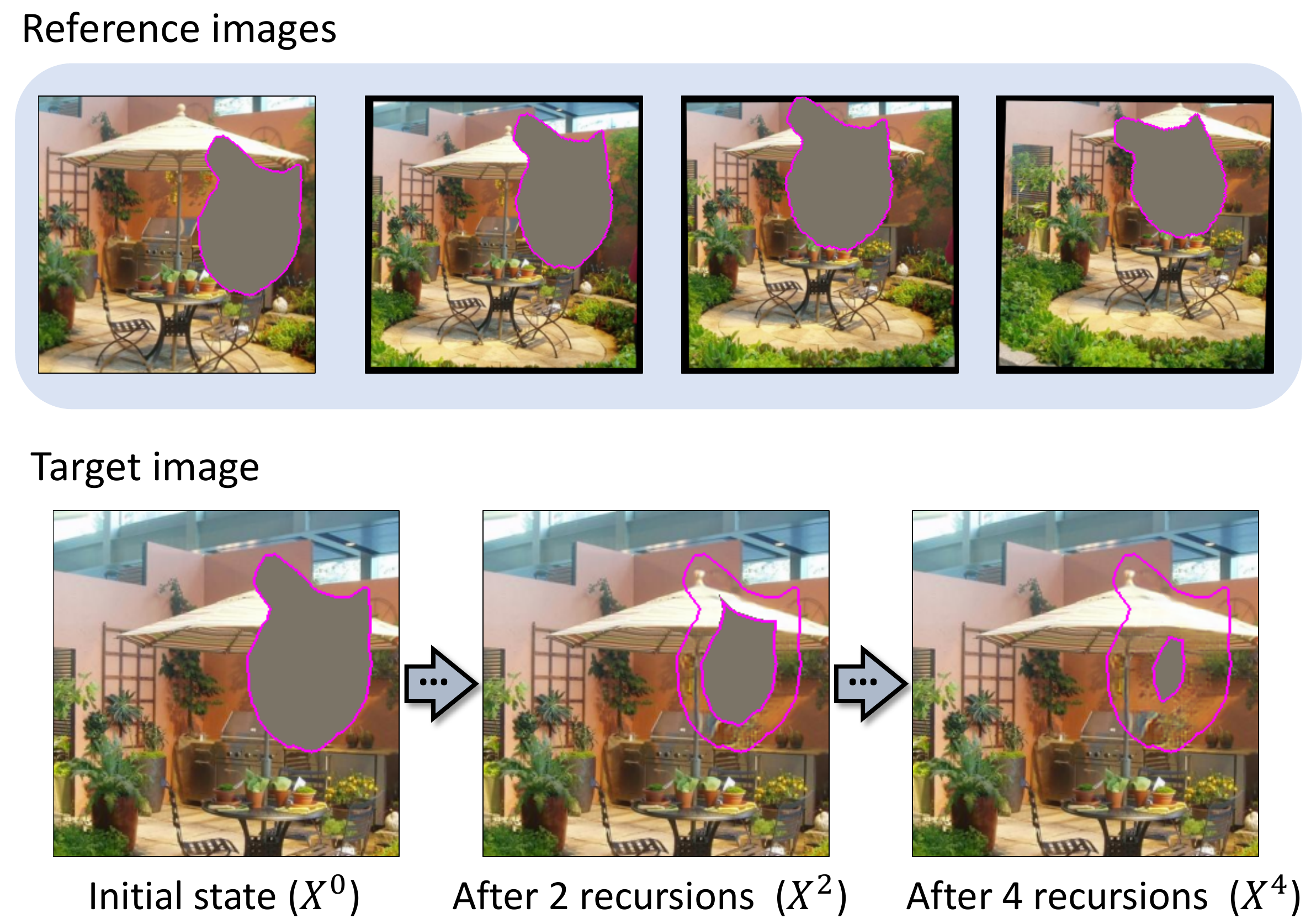}
\caption{Example of the training sample and the intermediate outputs. All the intermediate outputs are used for computing losses.}
\label{Fig:training}
\end{figure}

\subsection{Training}
\paragraph{Synthetic Training Data.}
Our framework is designed to retrieve the missing information from spatially and temporally distant pixels on the reference images.
This design choice allows us to easily synthesize the training data.
For each training sample, we prepare one target and four reference images taken from the same scene.
The network inputs are images with holes made using masks, and the network is trained to reconstruct the image pixels of the hole region in the target image. 

To synthesize such training samples, two ingredients are required: a set of images and masks. 
For images, we used two sources: Places2 dataset~\cite{zhou2017places} and YouTube videos that we collected ourselves.
Places2 dataset~\cite{zhou2017places} consists of 1.8 million images from 365 scene categories. 
With this dataset, we synthesized 5 background images by applying random affine transforms to an image such as rotation, translation, and cropping. 
The variety of contents in this dataset is useful for learning the generalization for various types of scenes.

In addition, we collected YouTube videos to learn a realistic scene transition between video frames. 
A total of 351 videos were collected from YouTube by searching for keywords like cities, nature, and world. 
Then, each video is chopped into short clips by detecting scene changes resulting in 7000 video clips.
During the training, 5 frames are randomly sampled from each video clip. 
The source of training data is randomly picked at an equal chance. 

For masks, we collected various object-shaped masks from the image datasets annotated with object masks (instance segmentation~\cite{everingham2010pascal, hariharan2011semantic} and salient object detection~\cite{shi2016hierarchical, cheng2015global}).
Then, these masks are randomly moved and deformed by random affine transforms to simulate moving objects. 
An example of our training data is shown in~\fref{Fig:training}.

\begin{figure*}
\centering
\includegraphics[width=1.0\linewidth]{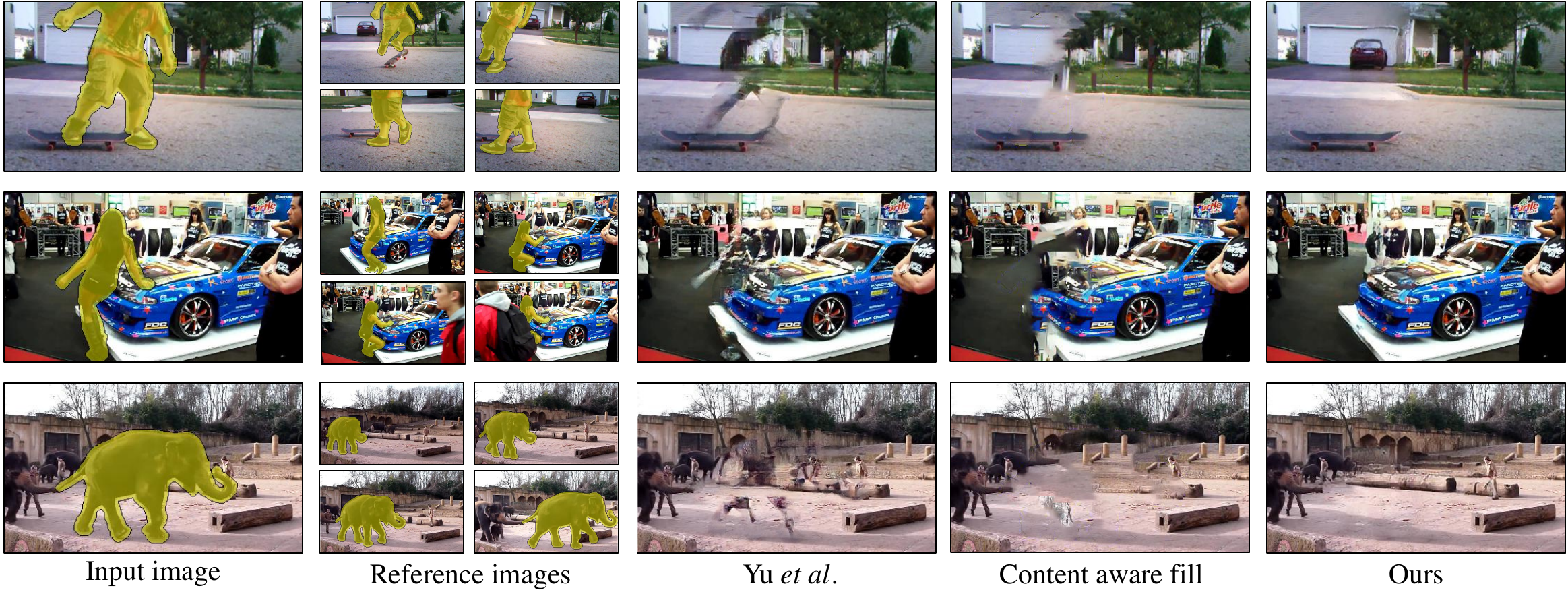}
\caption{Examples of image completion using a group of photos. The images are from Youtube-VOS~\cite{xu2018youtube_tech}.}
\label{Fig:group}
\end{figure*}


\paragraph{Loss Functions.}\label{Loss}
The loss function is designed to capture both the per-pixel reconstruction accuracy and the perceptual similarity.
To achieve this, we minimize the L1 distance to the groundtruth in the pixel space and the deep feature space at each recurrence step. 
Example intermediate outputs are shown in~\fref{Fig:training}.

The pixel losses are defined as follows: 
\begin{equation}
\begin{aligned}
    \mathcal{L}_{\text{peel}} &= \sum_{j}\|P^j\odot(\hat{X}^j - Y)\|_1,\\
    \mathcal{L}_{\text{valid}} &= \sum_{j}\|V\odot(\hat{X}^j - Y)\|_1,
\end{aligned}
\end{equation}
where $j$ indexes over the number of the recurrence, $\hat{X}^j$ is the raw output of the decoder before the composition at $j$-th recursion, $Y$ is the groundtruth, and $\odot$ indicates the element-wise multiplication.

Similarly, the perceptual loss is defined as the combination of the content and the style loss as follows:
\begin{equation}
\begin{aligned}
    \mathcal{L}_{\text{content}} &= \sum_{s=1}^3\sum_{j}\|\phi_{s}(X^j) - \phi_{s}(Y)\|_1, \\
    \mathcal{L}_{\text{style}} &= \sum_{s=1}^3\sum_{j}\|\text{G}(\phi_{s}(X^j)) - \text{G}(\phi_{s}(Y))\|_1,
\end{aligned}
\end{equation}
where $\phi_{s}(\cdot)$ is the mapping to $s$-th pooled feature map of VGG-16 network~\cite{simonyan2014very} pre-trained on ImageNet, and $\text{G}(\cdot)$ is a function for computing Gram matrix~\cite{gatys2016image}.

We compute the pixel losses directly on the decoder output and the perceptual loss after merging the peel pixels of the output with the non-hole region input.
The total loss is the weighted summation of all the loss functions.
\begin{equation}
\begin{aligned}
    \mathcal{L}_{\text{total}} &= 100\cdot\mathcal{L}_{\text{peel}} + \mathcal{L}_{\text{valid}} \\ & + 0.05\cdot\mathcal{L}_{\text{content}} + 120\cdot\mathcal{L}_{\text{style}} + 0.01\cdot\mathcal{L}_{\text{tv}},
\end{aligned}
\end{equation}
where $\mathcal{L}_{\text{tv}}$ is the total variation regularization term~\cite{johnson2016perceptual}. 
The balancing hyperparameters are adopted from ~\cite{Liu2018PartialConv}.

\paragraph{Training Details.}
We used 256$\times$256 images for the training and set the maximum number of the recursion as 5.  
For all experiments, the mini-batch size is set to 4 per GPU and Adam optimizer~\cite{kingma2014adam} is used.
The learning rate starts with 1e-4 and divided by 10 every 100,000 iteration. 
The training takes about 5 days using four NVIDIA V100 GPUs.

\subsection{Video Temporal Consistency}\label{TSN}
In the case of video completion, a video is processed frame-by-frame and the output videos often show flickering artifacts.
To remedy this, we post-process our video outputs with an additional temporal consistency networks.
The network is designed and trained for stabilizing videos inspired by~\cite{Lai-ECCV-2018} .
Specifically, an encoder-decoder network equipped with a convolutional GRU at the core is trained to balance between the temporal stability with the previous frame and the perceptual similarity with the current frame.
We modified the original method to match our need which is to stabilize the inpainted contents. 
Further details are covered in the supplementary materials and the effect of this post-processing is discussed in~\sref{Sect:pp}.

\begin{figure*}
\centering
\includegraphics[width=0.98\linewidth]{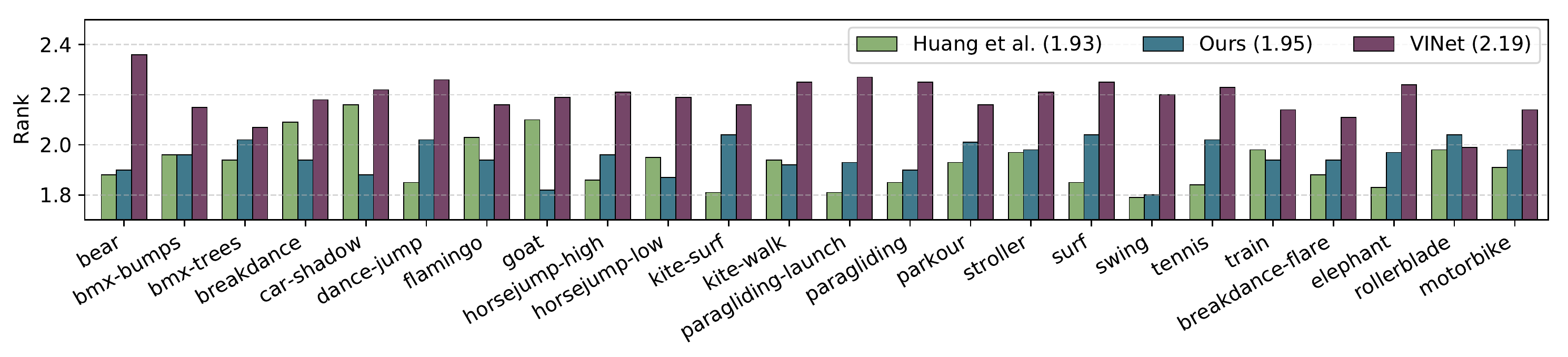}
\vspace{-4mm}
\caption{Result of our user study. The averaged ranks are shown in the figure legend.}
\vspace{-5mm}
\label{Fig:userstudy}
\end{figure*}

\vspace{-3pt}
\section{Experiments}
\vspace{-2pt}
In this section, we present our results on two experimental settings: reference-guided image completion and video completion. 

We conducted our experiments using the test images/videos with a resolution of 424$\times$240.
Our code and model will be available online.N
To obtain the results of other methods, we either used the available official codes or asked the authors for their results\footnote{\urlstyle{same} Yu~\etal~\cite{yu2018generative}: \url{https://github.com/JiahuiYu/generative_inpainting}, Huang~\etal~\cite{Huang2016siggraph}: \url{https://filebox.ece.vt.edu/~jbhuang/project/vidcomp/index.html}, and VINet~\cite{kim2019deep}: requested to the authors.}.





\subsection{Reference-guided Image Completion}
\vspace{-3pt}
In this setting, an algorithm aims to reconstruct the hole regions on an image using several reference images where the missing parts are partially visible. 

\paragraph{Test Data.}
There are few related inpainting methods \cite{wang2008stereoscopic,morse2012patchmatch,darabi2012image,barnes2015patchtable} that leverage reference views, but those are either limited to specific configurations or patch-based methods.
To the best of our knowledge, there is no deep learning method and standard test data specifically targeting this task. 
Therefore, we built a test set that consists of groups of photos, with each group having a set of photos of the same scene but in a different camera angle and time.
Each photo also needs to be annotated with the undesired objects.

Rather than collecting images that meet the requirements from scratch, we start from a video dataset labeled with objects. 
We take Youtube-VOS~\cite{xu2018youtube_tech} as the starting point.
From there, we manually select videos that contain the appropriate scenes, and 5 frames are uniformly sampled from each video to form a group of photos. 
Among 5 images, one is used as the target image and other four images are used as the reference images.

\paragraph{Comparison.}
We compare our method against the state-of-the-art learning-based~\cite{yu2018generative} and example-based~\cite{barnes2009patchmatch} completion methods. 

\noindent-- Yu~\etal~\cite{yu2018generative} is a generative model for single-image completion built upon the generative adversarial network~\cite{goodfellow2014generative}.
We used a model trained on Places2 data~\cite{zhou2017places}.

\noindent-- Content aware fill~\cite{barnes2009patchmatch}, a feature built into Adobe Photoshop, is the most popular tool for the image completion. 
To allow it to take advantages of the reference frames, we first concatenated the target and reference images, then we let it fill all the holes using all the available cues. 
 
As shown in~\fref{Fig:group}, the previous methods have difficulties in inpainting a large missing area and show artifacts. 
In comparison, our method is able to deal with a large hole through the onion peeling.
As expected, the single-image based method (Yu~\etal) tends to hallucinate contents, and inpainted results are not consistent with the reference frames. 
The example-based method (content aware fill) also fails to deliver pleasing results as it sometimes pastes wrong contents due to errors in the patch matching.
On the other hand, our method is able to fill holes with the actual contents by exploiting the reference images.
In the top row of ~\fref{Fig:group}, the red car completely occluded by a boy is recovered. 


\begin{figure*}
\centering
\includegraphics[width=1.0\linewidth]{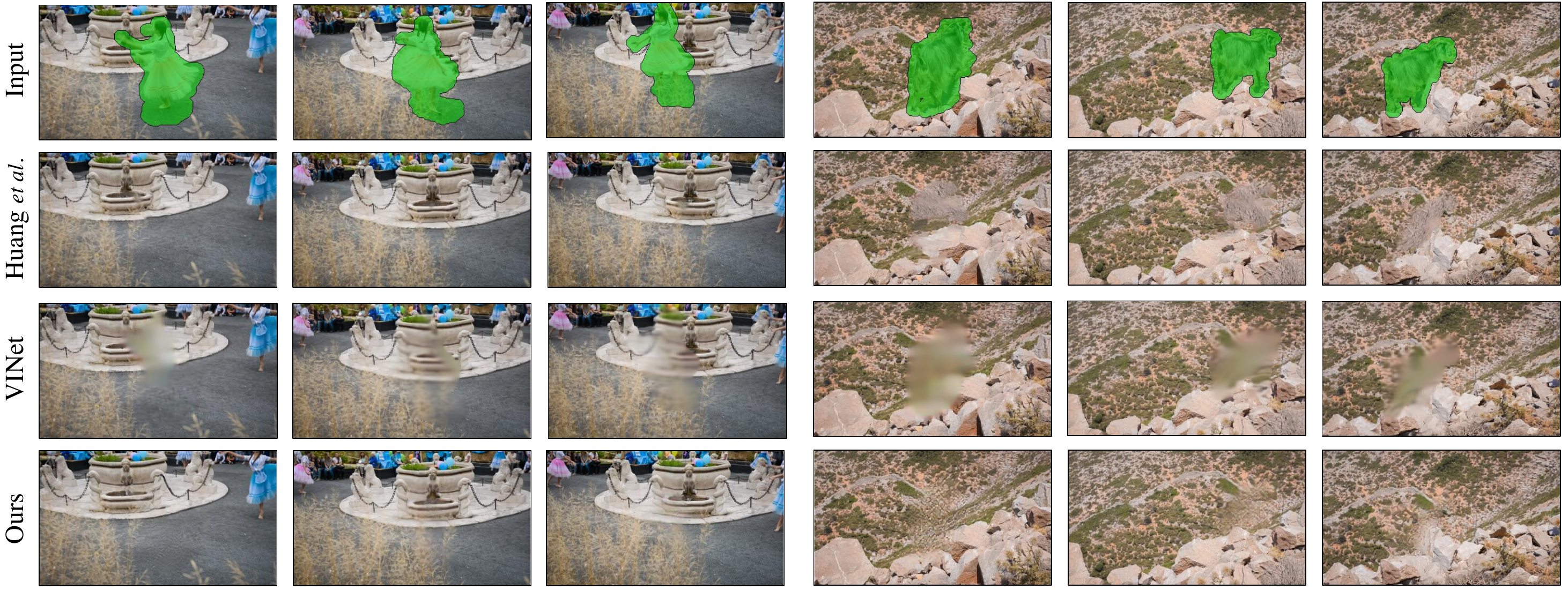}
\vspace{-18pt}
\caption{Visual comparison with video completion methods \cite{Huang2016siggraph,kim2019deep} on DAVIS~\cite{Perazzi2016}.}
\vspace{-10pt}
\label{Fig:vs_all}
\end{figure*}

\vspace{-2pt}
\subsection{Video Completion}
\vspace{-3pt}
In this section, our method is evaluated on the video completion task both quantitatively and qualitatively. 

\paragraph{User Study.}
We conducted a user study to subjectively compare our method against the state-of-the-art optimization-based~\cite{Huang2016siggraph} and learning-based~\cite{kim2019deep} video completion methods.



For the test videos, we used videos from DAVIS~\cite{Perazzi2016}, where every video frame has pixel-wise annotation for an object. 
We used 24 videos from DAVIS with additional annotations that include the shadow regions provided by~\cite{Huang2016siggraph}.

We conducted the user study through Amazon Mechanical Turk. 
For each video, the input and results from the 3 methods were shown to the participants, and the participants were asked to rank the results from 1 to 3. 
We allowed the participants to give a tie. 
Each test video was evaluated by 100 participants, and the averaged rank was as follows:

\begin{center}
\vspace{-3pt}
\begin{tabular}{c c P{2cm}}
\toprule
Huang~\etal~\cite{Huang2016siggraph}  & VINet~\cite{kim2019deep} & Ours\\ 
\midrule
1.93 & 2.19 & 1.95\\
\bottomrule
& & \small(lower is better) \\
\end{tabular}
\vspace{-4pt}
\end{center}

By the average rank, the quality of VINet~\cite{kim2019deep} is not on par with the other two methods.
The average ranks were very close between our method and \cite{Huang2016siggraph}, although \cite{Huang2016siggraph} had a slight edge. 
Note that our methods runs more than 50 times faster than \cite{Huang2016siggraph}. 
Per-video statistics are shown in~\fref{Fig:userstudy} and 
some sample results are shown in~\fref{Fig:vs_all}. 


\begin{figure}
\centering
\includegraphics[width=1.0\linewidth]{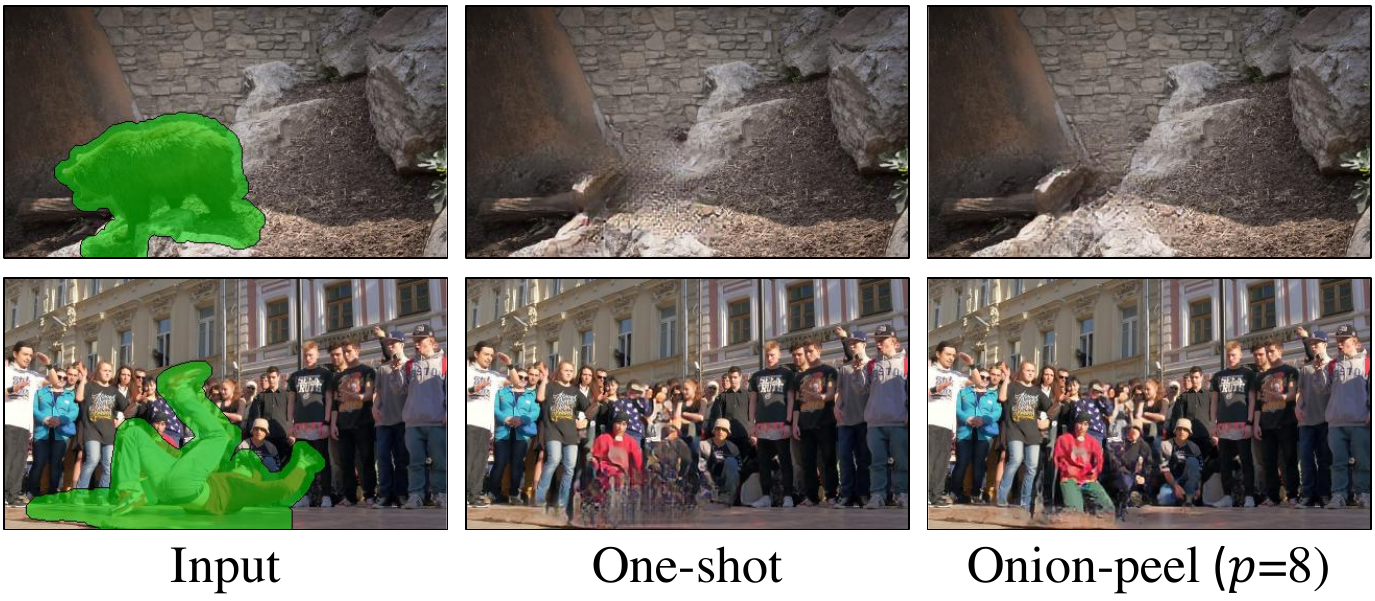}
\vspace{-20pt}
\caption{The effect of the onion peeling. }
\vspace{-5pt}
\label{Fig:peel}
\end{figure}

\begin{figure}
\centering
\includegraphics[width=1.0\linewidth]{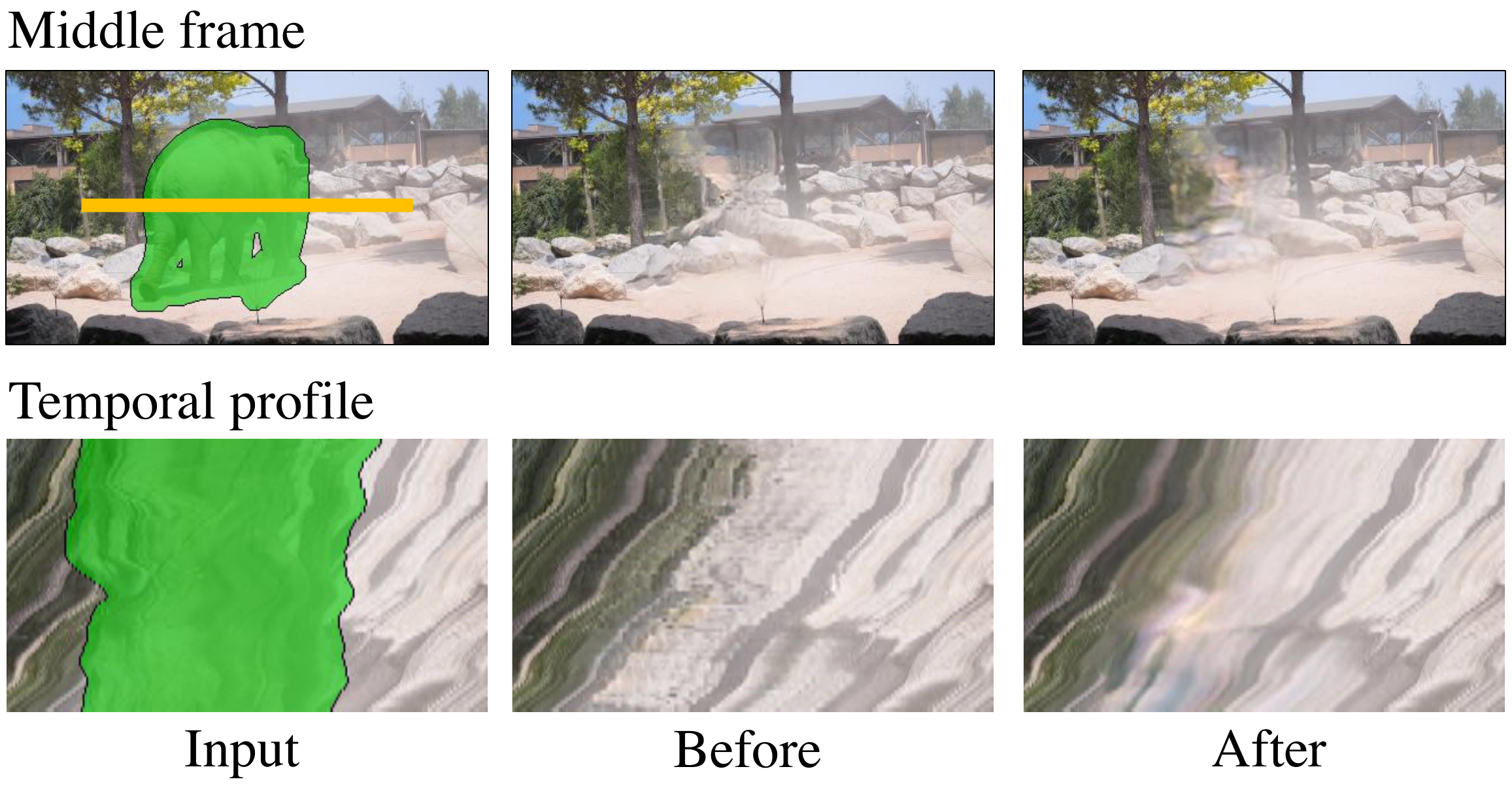}
\vspace{-20pt}
\caption{Our results before and after applying the post-processing. (top) the input and results at the middle frame. (bottom) temporal profile of the \textcolor{orange}{orange scan line}. The slices of video frames were stacked along the vertical-axis.}
\vspace{-10pt}
\label{Fig:vs_pp}
\end{figure}

\paragraph{Quantitative Evaluation.}
It is widely known that quantitatively evaluating the object removal is a hard task as the ground truth is not available. 
To make test videos with known backgrounds, we synthesized imaginary objects on top of the existing videos.
To achieve this, we shuffled the pairs of video and mask from DAVIS.
A total of 26 test videos were obtained this way, and used for comparing the pixel-wise reconstruction accuracy.  

In~\Tref{Table:psnr}, we quantitatively compare our method against the single-image~\cite{yu2018generative} and the video completion~\cite{Huang2016siggraph} method.
In addition to PSNR and SSIM, we provide Video Fr\'echet Inception Distance (VFID)~\cite{wang2018video} which is a video perceptual measure known to match well with human perception.
Our method shows the best accuracy on all the measures.
While the test videos are not real, we believe this synthetic experiments are helpful to understand the capability of different methods.


\vspace{-3pt}
\subsection{Analysis}\label{Sect:pp}
\vspace{-3pt}
\paragraph{Onion Peeling.} The core of our algorithm is the progressive image completion called the onion peeling. 
To validate its effect, we trained a variant of our model that inpaints the whole mask at one-shot.
The one-shot variant runs about 3$\times$ faster than our onion-peel network ($p$=8), but the output quality is not as good.
One-shot model often fails to handle large holes as shown in~\fref{Fig:peel}.

\begin{table}
\centering 
\begin{tabular}{lcccc}
\toprule
 & PSNR & SSIM & VFID & Time\\ 
\midrule
Yu~\etal~\cite{yu2018generative}  & 23.34 & 0.824 & 0.059&  \textbf{1.40}$s$\\
Huang~\etal~\cite{Huang2016siggraph} & 28.04 & 0.858 & 0.035 &  910$s$\\
Ours &  \textbf{30.19} &  \textbf{0.900} & \textbf{0.026} & 15.8$s$\\ 
\bottomrule
\end{tabular}
\vspace{-5pt}
\caption{Quantitative evaluation. The PSNR, SSIM and VFID are shown along with the execution time (sec/video). 
}
\label{Table:psnr}
\vspace{-15pt}
\end{table}

\paragraph{Post-processing.}
For video completion, we run post-processing for temporal consistency as described in~\sref{TSN}.
In~\fref{Fig:vs_pp}, we show the completion results before and after the post-processing. 
It clearly shows that the post-processing makes our results more temporally coherent as the temporal profiles are more smooth.
However, as a side effect, the temporal consistency network tends to blur video frames (\fref{Fig:vs_pp}).

\vspace{-3pt}
\section{Conclusion}
\vspace{-3pt}
We have presented a novel application of deep network for the image/video completion. 
For the video completion, our method shows a competitive quality to the state-of-the-art optimization based method while running in a fraction of time. 
Without any modification, our network is applicable to the image completion guided by the reference frames, which is hardly achieved by the existing methods. 
\\

\small{
\noindent\textbf{Acknowledgment.} This work was supported by the ICT R\&D program of MSIT/IITP (2014-0-00059) and the Technology Innovation Program (10073129) funded By the Ministry of Trade, industry \& Energy (MOTIE, Korea). 
}

{\small
\bibliographystyle{ieee_fullname}
\bibliography{egbib}

\begin{thebibliography}{10}\itemsep=-1pt

\bibitem{barnes2009patchmatch}
Connelly Barnes, Eli Shechtman, Adam Finkelstein, and Dan~B Goldman.
\newblock Patchmatch: a randomized correspondence algorithm for structural
  image editing.
\newblock {\em ACM Transactions on Graphics (TOG)}, 28(3):24, 2009.

\bibitem{barnes2015patchtable}
Connelly Barnes, Fang-Lue Zhang, Liming Lou, Xian Wu, and Shi-Min Hu.
\newblock Patchtable: Efficient patch queries for large datasets and
  applications.
\newblock {\em ACM Transactions on Graphics (TOG)}, 34(4):97, 2015.

\bibitem{cheng2015global}
Ming-Ming Cheng, Niloy~J Mitra, Xiaolei Huang, Philip~HS Torr, and Shi-Min Hu.
\newblock Global contrast based salient region detection.
\newblock {\em IEEE Transactions on Pattern Analysis and Machine Intelligence},
  37(3):569--582, 2015.

\bibitem{criminisi2004region}
Antonio Criminisi, Patrick P{\'e}rez, and Kentaro Toyama.
\newblock Region filling and object removal by exemplar-based image inpainting.
\newblock {\em IEEE Transactions on image processing}, 13(9):1200--1212, 2004.

\bibitem{darabi2012image}
Soheil Darabi, Eli Shechtman, Connelly Barnes, Dan~B Goldman, and Pradeep Sen.
\newblock Image melding: Combining inconsistent images using patch-based
  synthesis.
\newblock {\em ACM Transactions on Graphics (TOG)}, 31(4):82--1, 2012.

\bibitem{everingham2010pascal}
Mark Everingham, Luc Van~Gool, Christopher~KI Williams, John Winn, and Andrew
  Zisserman.
\newblock The pascal visual object classes (voc) challenge.
\newblock {\em International Journal of Computer Vision}, 88(2):303--338, 2010.

\bibitem{gatys2016image}
Leon~A Gatys, Alexander~S Ecker, and Matthias Bethge.
\newblock Image style transfer using convolutional neural networks.
\newblock In {\em Proceedings of the IEEE Conference on Computer Vision and
  Pattern Recognition}, pages 2414--2423, 2016.

\bibitem{goodfellow2014generative}
Ian Goodfellow, Jean Pouget-Abadie, Mehdi Mirza, Bing Xu, David Warde-Farley,
  Sherjil Ozair, Aaron Courville, and Yoshua Bengio.
\newblock Generative adversarial nets.
\newblock In {\em Advances in neural information processing systems}, pages
  2672--2680, 2014.

\bibitem{guillemot2014image}
Christine Guillemot and Olivier Le~Meur.
\newblock Image inpainting: Overview and recent advances.
\newblock {\em IEEE signal processing magazine}, 31(1):127--144, 2014.

\bibitem{hariharan2011semantic}
Bharath Hariharan, Pablo Arbel{\'a}ez, Lubomir Bourdev, Subhransu Maji, and
  Jitendra Malik.
\newblock Semantic contours from inverse detectors.
\newblock In {\em IEEE International Conference on Computer Vision (ICCV)},
  pages 991--998. IEEE, 2011.

\bibitem{hays2007scene}
James Hays and Alexei~A Efros.
\newblock Scene completion using millions of photographs.
\newblock {\em ACM Transactions on Graphics (TOG)}, 26(3):4, 2007.

\bibitem{huang2014image}
Jia-Bin Huang, Sing~Bing Kang, Narendra Ahuja, and Johannes Kopf.
\newblock Image completion using planar structure guidance.
\newblock {\em ACM Transactions on graphics (TOG)}, 33(4):129, 2014.

\bibitem{Huang2016siggraph}
Jia-Bin Huang, Sing~Bing Kang, Narendra Ahuja, and Johannes Kopf.
\newblock Temporally coherent completion of dynamic video.
\newblock {\em ACM Transactions on Graphics (TOG)}, 35(6), 2016.

\bibitem{IizukaSIGGRAPH2017}
Satoshi Iizuka, Edgar Simo-Serra, and Hiroshi Ishikawa.
\newblock {Globally and Locally Consistent Image Completion}.
\newblock {\em ACM Transactions on Graphics (Proc. of SIGGRAPH 2017)},
  36(4):107:1--107:14, 2017.

\bibitem{ilan2015survey}
Shachar Ilan and Ariel Shamir.
\newblock A survey on data-driven video completion.
\newblock In {\em Computer Graphics Forum}, volume~34, pages 60--85. Wiley
  Online Library, 2015.

\bibitem{johnson2016perceptual}
Justin Johnson, Alexandre Alahi, and Li Fei-Fei.
\newblock Perceptual losses for real-time style transfer and super-resolution.
\newblock In {\em European conference on computer vision}, pages 694--711.
  Springer, 2016.

\bibitem{kim2019deep}
Dahun Kim, Sanghyun Woo, Joon-Young Lee, and In~So Kweon.
\newblock Deep video inpainting.
\newblock In {\em Proceedings of the IEEE Conference on Computer Vision and
  Pattern Recognition}, 2019.

\bibitem{kingma2014adam}
Diederik Kingma and Jimmy Ba.
\newblock Adam: A method for stochastic optimization.
\newblock In {\em International Conference on Learning Representations}, 2015.

\bibitem{Lai-ECCV-2018}
Wei-Sheng Lai, Jia-Bin Huang, Oliver Wang, Eli Shechtman, Ersin Yumer, and
  Ming-Hsuan Yang.
\newblock Learning blind video temporal consistency.
\newblock In {\em European Conference on Computer Vision}, 2018.

\bibitem{Liu2018PartialConv}
Guilin Liu, Fitsum A.~Reda, Kevin J.~Shih, Ting-Chun Wang, and Andrew Tao~Bryan
  Catanzaro.
\newblock Image inpainting for irregular holes using partial convolutions.
\newblock In {\em IEEE International Conference on Computer Vision (ICCV)},
  2018.

\bibitem{liu2013exemplar}
Yunqiang Liu and Vicent Caselles.
\newblock Exemplar-based image inpainting using multiscale graph cuts.
\newblock {\em IEEE transactions on image processing}, 22(5):1699--1711, 2013.

\bibitem{morse2012patchmatch}
Bryan Morse, Joel Howard, Scott Cohen, and Brian Price.
\newblock Patchmatch-based content completion of stereo image pairs.
\newblock In {\em 2012 Second International Conference on 3D Imaging, Modeling,
  Processing, Visualization \& Transmission}, pages 555--562. IEEE, 2012.

\bibitem{Newson2014siam}
Alasdair Newson, Andrés Almansa, Matthieu Fradet, Yann Gousseau, and Patrick
  Pérez.
\newblock Video inpainting of complex scenes.
\newblock {\em SIAM Journal on Imaging Sciences, Society for Industrial and
  Applied Mathematics}, 7(4):1993--2019, 2014.

\bibitem{newson2017non}
Alasdair Newson, Andr{\'e}s Almansa, Yann Gousseau, and Patrick P{\'e}rez.
\newblock Non-local patch-based image inpainting.
\newblock {\em Image Processing On Line}, 7:373--385, 2017.

\bibitem{PathakCVPR2016}
Deepak Pathak, Philipp Krahenbuhl, Jeff Donahue, Trevor Darrell, and Alexei
  A.~Efros.
\newblock Context encoders: Feature learning by inpainting.
\newblock In {\em IEEE Conference on Computer Vision and Pattern Recognition
  (CVPR)}, 2016.

\bibitem{Perazzi2016}
Federico Perazzi, Jordi Pont-Tuset, Brian McWilliams, Luc Van~Gool, Markus
  Gross, and Alexander Sorkine-Hornung.
\newblock A benchmark dataset and evaluation methodology for video object
  segmentation.
\newblock In {\em IEEE Conference on Computer Vision and Pattern Recognition
  (CVPR)}, 2016.

\bibitem{shi2016hierarchical}
Jianping Shi, Qiong Yan, Li Xu, and Jiaya Jia.
\newblock Hierarchical image saliency detection on extended cssd.
\newblock {\em IEEE Transactions on Pattern Analysis and Machine Intelligence},
  38(4):717--729, 2016.

\bibitem{simonyan2014very}
Karen Simonyan and Andrew Zisserman.
\newblock Very deep convolutional networks for large-scale image recognition.
\newblock {\em arXiv preprint arXiv:1409.1556}, 2014.

\bibitem{sukhbaatar2015end}
Sainbayar Sukhbaatar, Jason Weston, Rob Fergus, et~al.
\newblock End-to-end memory networks.
\newblock In {\em Advances in neural information processing systems}, pages
  2440--2448, 2015.

\bibitem{wang2018videoinp}
Chuan Wang, Haibin Huang, Xiaoguang Han, and Jue Wang.
\newblock Video inpainting by jointly learning temporal structure and spatial
  details.
\newblock In {\em Proceedings of the 33th AAAI Conference on Artificial
  Intelligence}, 2019.

\bibitem{wang2008stereoscopic}
Liang Wang, Hailin Jin, Ruigang Yang, and Minglun Gong.
\newblock Stereoscopic inpainting: Joint color and depth completion from stereo
  images.
\newblock In {\em 2008 IEEE Conference on Computer Vision and Pattern
  Recognition}, pages 1--8. IEEE, 2008.

\bibitem{wang2018video}
Ting-Chun Wang, Ming-Yu Liu, Jun-Yan Zhu, Guilin Liu, Andrew Tao, Jan Kautz,
  and Bryan Catanzaro.
\newblock Video-to-video synthesis.
\newblock In {\em Advances in neural information processing systems}, 2018.

\bibitem{wang2018non}
Xiaolong Wang, Ross Girshick, Abhinav Gupta, and Kaiming He.
\newblock Non-local neural networks.
\newblock In {\em IEEE Conference on Computer Vision and Pattern Recognition
  (CVPR)}, 2018.

\bibitem{wexler2004space}
Yonatan Wexler, Eli Shechtman, and Michal Irani.
\newblock Space-time video completion.
\newblock In {\em Proceedings of IEEE Computer Society Conference on Computer
  Vision and Pattern Recognition}, 2004.

\bibitem{xu2018youtube_tech}
Ning Xu, Linjie Yang, Dingcheng Yue, Jianchao Yang, Yuchen Fan, Yuchen Liang,
  and Thomas Huang.
\newblock Youtube-vos: A large-scale video object segmentation benchmark.
\newblock In {\em arXiv preprint arXiv:1809.03327}, 2018.

\bibitem{yu2018free}
Jiahui Yu, Zhe Lin, Jimei Yang, Xiaohui Shen, Xin Lu, and Thomas~S Huang.
\newblock Free-form image inpainting with gated convolution.
\newblock {\em arXiv preprint arXiv:1806.03589}, 2018.

\bibitem{yu2018generative}
Jiahui Yu, Zhe Lin, Jimei Yang, Xiaohui Shen, Xin Lu, and Thomas~S Huang.
\newblock Generative image inpainting with contextual attention.
\newblock In {\em Proceedings of the IEEE Conference on Computer Vision and
  Pattern Recognition}, pages 5505--5514, 2018.

\bibitem{zhou2017places}
Bolei Zhou, Agata Lapedriza, Aditya Khosla, Aude Oliva, and Antonio Torralba.
\newblock Places: A 10 million image database for scene recognition.
\newblock {\em IEEE Transactions on Pattern Analysis and Machine Intelligence},
  2017.

\end{thebibliography}
}

\end{document}